\documentclass[10pt,twocolumn,letterpaper]{article}

\usepackage{wacv}
\usepackage{times}
\usepackage{epsfig}
\usepackage{graphicx}
\usepackage{amsmath}
\usepackage{amssymb}
\usepackage{floatrow}
\usepackage{subfig}
\usepackage{color}
\usepackage{textpos}

\usepackage[pagebackref=true,breaklinks=true,letterpaper=true,colorlinks,bookmarks=false]{hyperref}
\def\equationautorefname~#1\null{%
  (#1)\null}
\def\subsectionautorefname~#1\null{%
  #1\null}

\renewcommand{\subsectionautorefname}{section}

\wacvfinalcopy 

\def\wacvPaperID{472} 

\ifwacvfinal\fi
\begin{document}

\title{Automatic Analysis of Sewer Pipes Based on Unrolled Monocular Fisheye Images}


\author{
\makebox[.4\linewidth]{Johannes K\"unzel} \\ Humboldt University Berlin\\
{\tt\small johannes.kuenzel@hu-berlin.de}
\and
\makebox[.4\linewidth]{Thomas Werner} \\ Fraunhofer IAIS\\
{\tt\small thomas.werner@iais.fraunhofer.de}
\and
\makebox[.4\linewidth]{Ronja M\"oller} \\ Fraunhofer IAIS\\
{\tt\small ronja.moeller@iais.fraunhofer.de}
\and
\makebox[.4\linewidth]{Peter Eisert} \\ Humboldt University Berlin\\
{\tt\small peter.eisert@hu-berlin.de}
\and
\makebox[.4\linewidth]{Jan Waschnewski} \\ Berliner Wasserbetriebe \\
{\tt\small jan.waschnewski@bwb.de}
\and
\makebox[.4\linewidth]{Ralf Hilpert} \\ Berliner Wasserbetriebe \\
{\tt\small ralf.hilpert@bwb.de}
}

\maketitle
\ifwacvfinal\thispagestyle{empty}\fi

\begin{abstract}
The task of detecting and classifying damages in sewer pipes offers an important application area for computer vision algorithms.
This paper describes a system, which is capable of accomplishing this task solely based on low quality and severely compressed fisheye images from a pipe inspection robot.
Relying on robust image features, we estimate camera poses, model the image lighting, and exploit this information to generate high quality cylindrical unwraps of the pipes' surfaces.
Based on the generated images, we apply semantic labeling based on deep convolutional neural networks to detect and classify defects as well as structural elements.
\end{abstract}

\begin{textblock*}{175mm}(0mm,85mm)
\scriptsize{\copyright 2018 IEEE. Published in the IEEE 2018 Winter Conference on Applications of Computer Vision (WACV 2018), scheduled for 12-14 March 2018 in Lake Tahoe, NV/CA. Personal use of this material is permitted. However, permission to reprint/republish this material for advertising or promotional purposes or for creating new collective works for resale or redistribution to servers or lists, or to reuse any copyrighted component of this work in other works, must be obtained from the IEEE.\\}
\end{textblock*}
\begin{textblock*}{175mm}(0mm,94mm)
\scriptsize{The final published version is available at https://doi.org/10.1109/WACV.2018.00223}
\end{textblock*}

\section*{Acknowledgment}
This work is funded by the German Federal Ministry of Education and Research under grant number 13N13891.

\section{Introduction}
\label{sec:intro}

The inspection of sewer pipes is a crucial task to ensure the functionality of sewage systems.
Many sewer pipes in big cities are several decades old, some are even older than one hundred years. 
Therefore, regular risk assessment and sanitation planing is needed to ensure the correct functionality of the sewer system.
At present, mobile robot systems equipped with cameras or other sensors are used to manually traverse the pipes.
As a result, they produce large amounts of data in which defects have to be annotated manually by technical staff especially trained for this task. 
As a consequence, the obtained results are error prone due to the repeating and tiresome work. 
In order to assist workers, reliable computer systems are needed that can give support by automatically detecting certain defects in sewer pipes. 
Such a system would typically consist of to modules: First, a pre-processing module that can convert raw input images into a form that can be automatically processed, and second, a detection/classification module that performs automated annotation of the provided data.

One early work for the estimation of camera poses in sewer papers was proposed by Cooper \etal.~\cite{Cooper.1998}.
The authors exploited the longitudinal mortar lines for camera pose recovery, limiting the system to stonewalled pipes.
In \cite{Kannala.2008}, the profile of sewer pipes is reconstructed solely from fisheye video sequences.
The approach is based on the tracking of feature points for more than three views, which is not feasible in our application due to the distance of 5 cm and thus large changes between consecutive images.
Furthermore, the system was only tested for concrete pipes, which have a relatively well structured texture for feature detection and matching.
Esquivel \etal.~\cite{Esquivel.2005,Esquivel.2010} proposed a system for the reconstruction of sewer shafts exploiting the fact that the camera always faces downward due to the force of gravity.
Therefore, the system is restricted to vertical pipes.

With our work, we present a system, which assists the employee with the automatic detection and classification of damages in sewer pipes.
We solely use unrolled and stitched images (like the one in \autoref{fig:comparsion}) as input for the detection and classification algorithm.
To obtain an unrolled fisheye image, 3d motion of the camera is tracked and a cylindrical image is generated through back-projection on an ideal pipe.
This image can then be easily snipped and unwound into a planar image.

The second part of this work is aimed towards automatic detection and classification of defects and structural elements in the pipe.
We treat this as a semantic labeling problem which is tackled using deep convolutional neural networks. To our knowledge previous methods for this task mostly relied on image processing algorithms and heuristics to detect various types of defects and structural elements. 
In \cite{Su2015} for example, the authors use edge detection and morphological operations based on CCTV images for crack and open joint detection.
The authors of \cite{Huynh2015} propose a novel edge detection algorithm for thin crack detection, that can overcome some difficulties encountered in noisy environments like sewer pipes.

Besides those algorithms, few works also use machine learning to implement diagnostic systems based on a range of image processing techniques. Yang and Su \cite{Yang2008} for example use SVMs and simple neural networks using wavelet transform and co-occurrence matrices to detect open joints, cracks and broken pipes. Another example is presented by Wu \etal in \cite{Wu2015} where ensemble methods on contourlet transforms and statistical features are used to detect cracks, roots and collapsed pipes.

\begin{figure}
\centering
\includegraphics[width=0.6\linewidth]{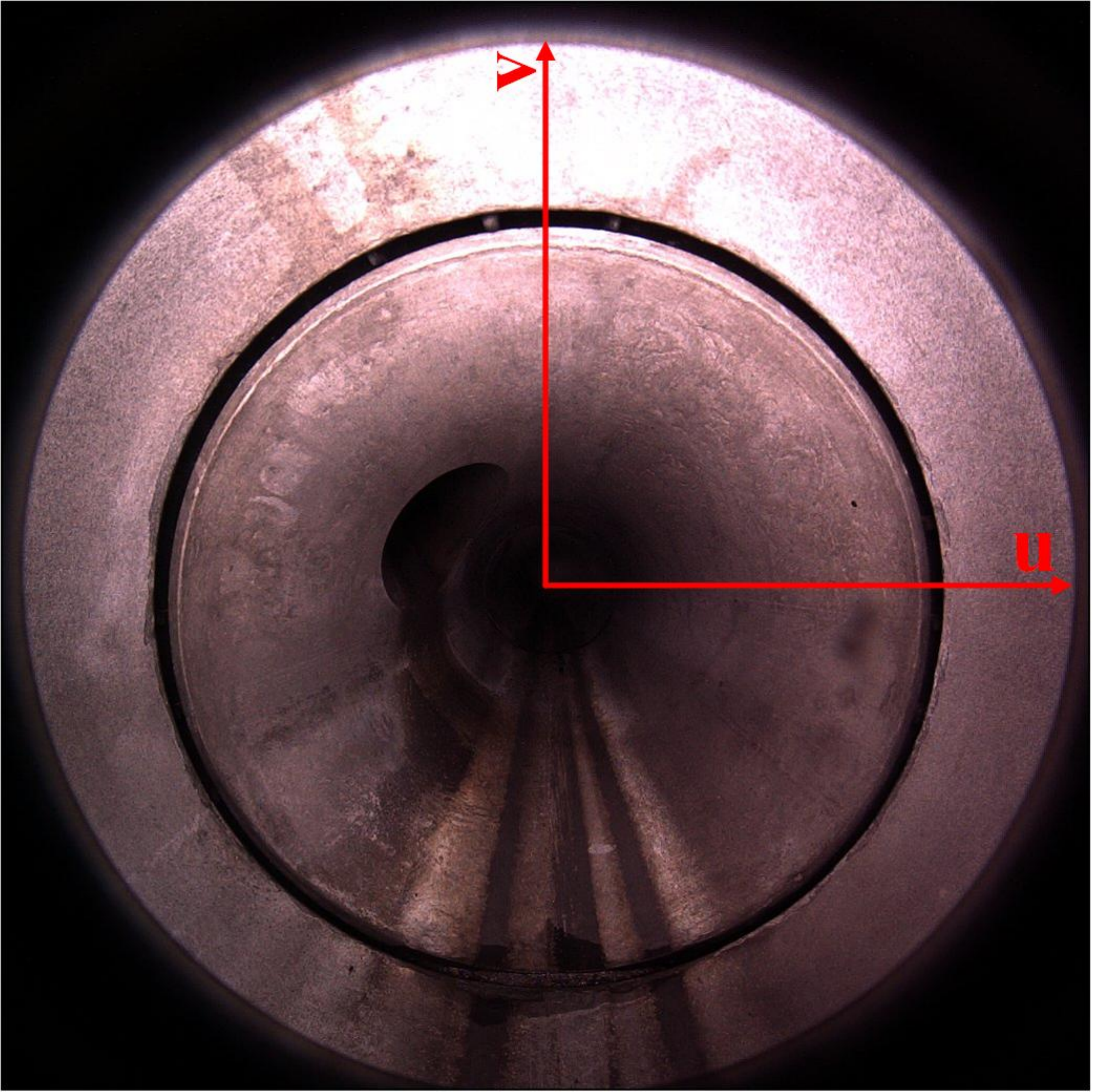}
\caption{Sample fisheye image taken with a mobile robot inside the pipe. The center of the round image area on the sensor is given.}
\label{fig:sampleFisheyeImage}
\end{figure}

\section{Creation of Unrolled Pipe Images}
\label{sec:imagePreProcessing}
The images of the sewer pipe are taken with a mobile robot having a fisheye camera with a field of view of 185 degrees.
Due to their characteristics and the viewing direction along the cylindrical pipe, the resolution of the depicted pipe surface decreases dramatically with increasing distance to the camera.
In consequence, only the outer part of the circular image is used to produce the stitched unroll of the sewer pipe.
To guarantee overlap big enough for registration, images are taken with five centimeter spacing.

The original images we obtain from the commercial robot system show strong artifacts from severe lossy compression and some image areas are overexposed because of the strong flashlights.
In conjunction with the lack of texture information, classical approaches used by systems like Bundler \cite{Snavely.2005} or VisualSfM \cite{Wu.2011,Wu.2013} fail to track our robot camera.
We therefore simplify the problem by assuming a cylindrical shape with a known diameter (for the absolute scale).
Instead of having one unknown depth parameter for each image feature, the 3d position of all features can now be related by one unknown rigid body transform with 6 unknowns for the entire frame.
The resulting equation system and its solution are explained in the following sections.

\subsection{Back-Projection}
\label{ssec:backProjection}

\begin{figure}[h]
\centering
\includegraphics[width=0.95\linewidth]{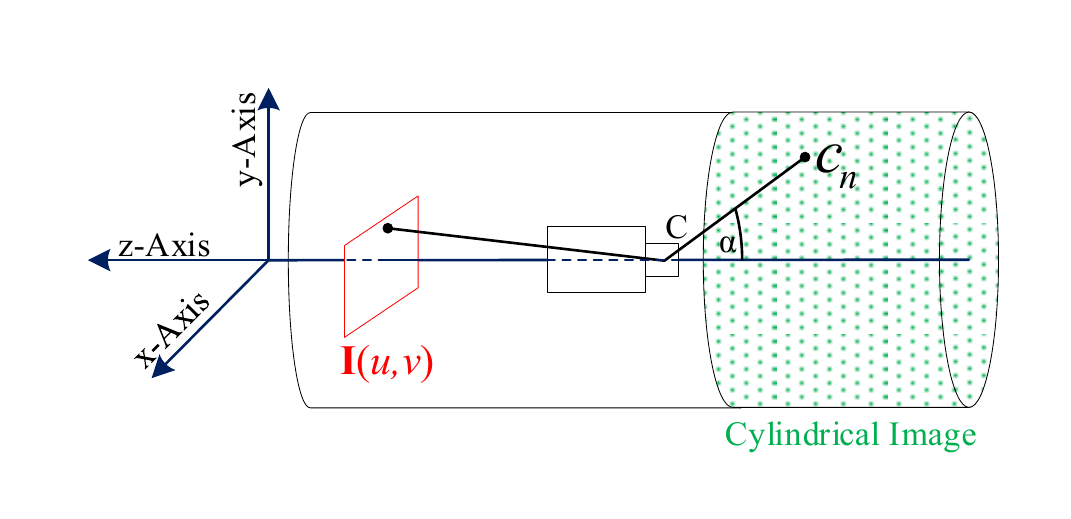}
\caption{Scheme illustrating the back-projection of the point $\mathbf{p}_n$ on the image plane, resulting in point $\mathbf{p}_n^\prime$. Interpolation is done at $\mathbf{p}_n^\prime$ to retrieve the color for $\mathbf{p}_n$.}
\label{fig:backProjectionScheme}
\end{figure}

The imaging characteristics of the circular fisheye lenses can be described by

\begin{equation}
\frac{\alpha}{d} = \frac{\Gamma}{D}
\label{eq:lens_charateristics}
\end{equation}

with $\alpha$ representing the angle of incidence, $d$ the distance of the resulting image point from the image center in pixel, $\Gamma$ the field of view of the fisheye lens and $D$ the diameter of the projected circular part of the image in pixel. 
For the unwrapping, we regulary sample the cylindrical surface of the inner pipe at N 3d positions $\mathbf{P}=\{\mathbf{p}_1,\ldots,\mathbf{p}_N\}$ which correspond to the pixels of the cylindrical image (please refer to \autoref{fig:backProjectionScheme} for an illustration).
One can choose a reasonable number of points lying on the pipe perimeter, depending on the resolution of the source images.
The spatial resolution of the pipe perimeter defines the resolution along the pipe axis as well if square pixels are assumed.
Every point $\mathbf{p}$ can then be projected into the corresponding fisheye image to acquire the color information at the location $\mathbf{I}(u,v)$.
Usually, the projection will hit the image at subpixel positions, therefore image interpolation is needed.

Subsequently, the cylindrical unwrap for every fisheye image can be calculated for a given camera pose.
In the existing commercial system, the camera is assumed to lie on and move along the pipe axis.
This assumption is not valid due to the movement between the shots -- major stitching artifacts are the consequence, making it nearly impossible to use these images for training of neural networks for automatic damage detection.

\subsection{Camera Pose Estimation}

We chose a feature based approach for the estimation of the camera pose.
In \cite{Zhang.2011}, features are detected and filtered based on the images generate by back-projection.
Due to the unknown camera pose, fairly strong artifacts may occur, causing less features or false matches. 
Due to the relatively homogeneous pipe surface and the limited image quality, we use an iterative feature matching scheme \cite{Furch.2013} that considers neighborhood constraints and leads to more and more robust feature correspondences.

\paragraph*{Local Pose Optimization}

Utilizing \autoref{eq:lens_charateristics}, describing the characteristics of the fisheye lens, a vector

\begin{equation}
\mathbf{d_i} = 
\begin{bmatrix}
u \\
v \\
\frac{d}{\tan(\alpha)}
\end{bmatrix}\textnormal{, with } (d > 0)
\end{equation}

can be constructed for every feature point $\mathit{F_i}(u,v)$ to describe the direction of the corresponding line of sight.
With respect to a global coordinate system placed on the symmetric axis of the pipe, the position and orientation of the robot camera can be specified by a translation vector $\mathbf{t}$ and a rotation matrix $\mathbf{R}$, respectively.
For this pose, the straight-line equation

\begin{equation}
\mathbf{g_i} = \mathbf{t} + \lambda \cdot \mathbf{R} \cdot \mathbf{d_i}\textnormal{, with } (\lambda \geq 0)
\label{eq:straightLine}
\end{equation}

represents the line of sight originating from the camera center, depending on the camera position.
Based on \autoref{eq:straightLine}, the intersections of the lines with the cylinder surface of the pipe can be calculated.
For matched feature points, the intersections must be consistent on the 3d surface.

Our objective is to combine the equations in a linear equation system, which can be solved efficiently for the translational ($\Delta\mathbf{t}$) and rotational ($\Delta\mathbf{R}$) updates.
We linearize \autoref{eq:straightLine} with respect to the rotation angles $\alpha_i$
\begin{equation}
\begin{aligned}
\mathbf{g} &= \mathbf{t} + \boldsymbol{\Delta t} + \lambda \cdot \boldsymbol{\Delta R} \cdot \mathbf{R} \cdot  \mathbf{d} \\
&= \underbrace{\mathbf{t} + 
\begin{bmatrix}
\Delta t_x \\
\Delta t_y \\
\Delta t_z
\end{bmatrix}}_{\mathbf{a}(\Delta t_x, \Delta t_y, \Delta t_z)}
+ \lambda \cdot
\underbrace{\begin{bmatrix}
1 & -\alpha_z & \alpha_y \\
\alpha_z & 1 & -\alpha_x \\
-\alpha_y & \alpha_x & 1
\end{bmatrix}
\cdot \mathbf{R} \cdot  \mathbf{d}}_{\mathbf{b}(\alpha_x, \alpha_y, \alpha_z)}
\end{aligned}
\label{eq:linearizedStraightLine}
\end{equation}
around an operating point of $\mathbf{t}$ and $\mathbf{R}$, resulting in three equations, one for each component of $\mathbf{g}$.
With the circle equation $x^2+y^2=r^2$ (with $r$ being the radius of the pipe) and the x- and y-component of the linearized straight-line equation \autoref{eq:linearizedStraightLine}, the function for $\lambda(\Delta t_x, \Delta t_y, \alpha_x, \alpha_y, \alpha_z)$ specifying the intersection point can be created.
So the location of a certain point $\mathbf{p}=[p_x  p_y  p_z]^T$ on the pipe surface can be specified by 
\begin{equation}
\begin{split}
p_x( &\Delta t_x, \Delta t_y, \alpha_x, \alpha_y, \alpha_z ) = a_x( \Delta t_x ) \\
& + \lambda(\Delta t_x,\Delta t_y,\alpha_x,\alpha_y,\alpha_z) \cdot b_x( \alpha_x, \alpha_y, \alpha_z )
\end{split}
\end{equation}
for the first dimension.
The equations for the remaining dimensions have the same structure but with  $a_y(\Delta t_y)$ and $a_z(\Delta t_z)$, respectively.

In our local pose estimation scheme, camera location and orientation is determined from point correspondences between two successive camera frames with unknown pose.
Therefore, we always estimate pairs of 3D camera data corresponding to the two frames. 
To avoid pose ambiguity of the cylindrical shape, the first camera of each pair has to be fixed in its position along the z-axis and in its rotation around it, resulting in $4+6=10$ unknown pose parameters.
Based on the partial derivatives for the parameters around an initial operating point, a linear equation system is created and solved for every camera pair.
This procedure is iteratively applied to remove errors due to linearization.

After each iteration, the parameter vector must be updated which results in a simple addition for the translation parameters.
The updated rotation matrix $\mathbf{R^*}$ is calculated by $\mathbf{R^*} = \mathbf{R}_{\alpha_x} \cdot \mathbf{R}_{\alpha_y} \cdot \mathbf{R}_{\alpha_z} \cdot \mathbf{R}$.
%
%

During the iterations, we utilize the RANSAC algorithm to reject outliers.
This makes the pose estimation more robust against e.g.~connecting pipes or dangling roots, which introduce many features violating the assumption of a cylindrically shaped pipe with a known diameter.
Otherwise, the pose estimation would likely fail at these points.

\paragraph*{Global Pose Optimization}

In the local optimization used for initialization, camera pose is estimated independently for each pair of frames.
To get a smooth camera path, a global optimization of the camera poses is done, since pose of the current frame directly depends on pose of the previous frame.
Every camera but the first has therefore six degrees of freedom and the pose is determined by the location of the matched features for both pairs the image is involved in.
The first frame only has four degrees of freedom due to the ambiguity mentioned above.
The matrices for $K$ frames are combined into one big sparse matrix with $6 \cdot K - 2$ unknowns for which the linear equation system is solved iteratively.
Therefore, all camera parameters are connected throughout the whole camera path and can influence each other during the optimization.
Due to the initialization with the results of the local estimation, this equation system can be solved quickly by exploiting its sparse nature.
With the estimated camera poses in place, the back-projection of the unwrapped image can be calculated without artifacts caused by unconsidered camera movement.

\subsection{Image Enhancement and Stitching}

\begin{figure}
\centering
\includegraphics[width=1.0\linewidth]{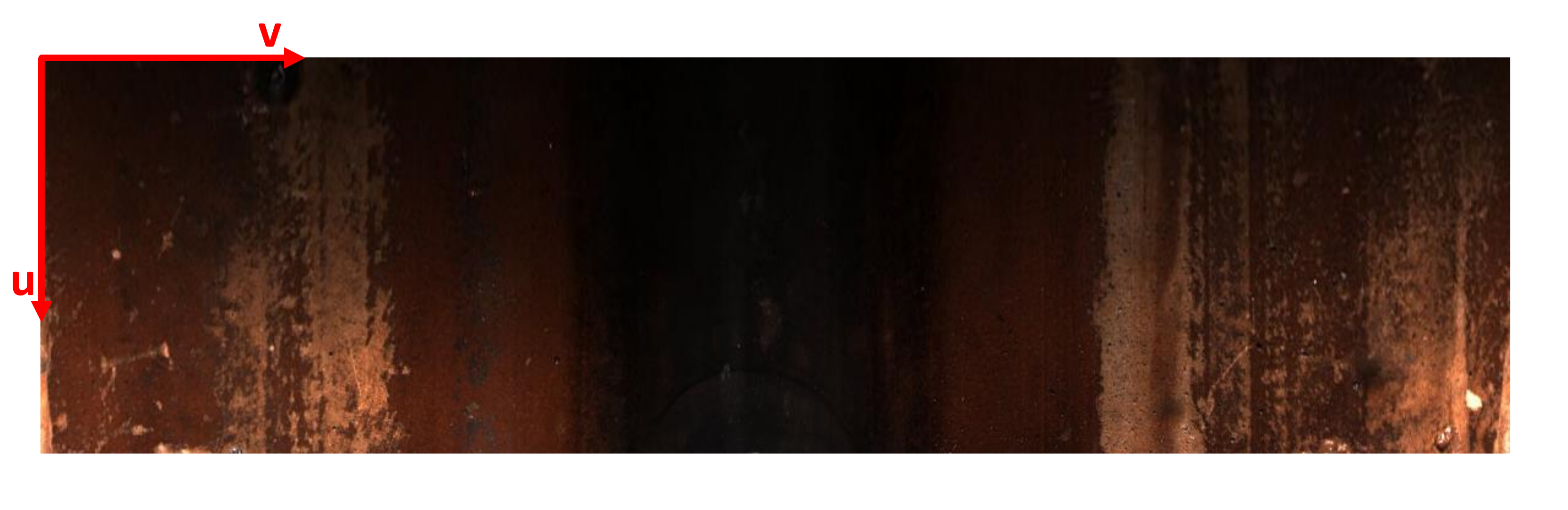}
\caption{Single unwrapped fisheye image after camera pose estimation. There is a clearly visible light falloff with increasing distance to the camera.}
\label{fig:sampleUnwrapImage}
\end{figure}

Despite the removal of geometric artifacts, caused by the camera motion, there are still major artifacts caused by the uneven lighting of the images.
The radial light falloff is easily noticeable in \autoref{fig:sampleFisheyeImage} and \autoref{fig:sampleUnwrapImage} causing a leap in lighting from dark to bright at the seam between two images.
To get a smooth imperceptible transition between consecutive images, we apply these three steps:
\begin{enumerate}
\item Estimation and elimination of uneven lighting.
\item Identification of the optimal seam using Dynamic Programming.
\item Application of Poisson Blending in the transition zone. 
\end{enumerate}

\paragraph*{Elimination of uneven lighting}
In contrast to \cite{Zhang.2011}, the illumination is modeled separately for each image.
The authors of \cite{Zhang.2011} assume that the average grey level of a distinct pixel can be regarded as the illumination intensity.
This assumption is only valid if the reflection properties and color remain the same over the entire length of the pipe.
In our use case though, we have to deal with changing materials, and deposits on the surface affecting the reflection of light.
Therefore, we modulate the lighting falloff for each image separately as a linear function of the distance to the camera.

The lighting estimation is done on the grey-scale images $\mathbf{G}(u,v)$ in several steps.
To separate the low frequencies, a fairly strong Gaussian filter is applied.
This practice helps to reduce the influence of locally strong reflections or dark spots, like pipe connections.
After that, a linear function $L_v(u)$ is fitted to every image column $v$.
We use a robust regression method by fitting a linear function iteratively to the image while trimming out image areas with the biggest residuals.
In every image column $v$, we also calculate the median value $O_v$ from all grey values to get an estimate for the offset.
We then adjust every channel of $\mathbf{I}$ with

\begin{equation}
\mathbf{I}^*(u,v) = \mathbf{I}(u,v) - L_v(u) + O_v 
\end{equation}
getting the new corrected image $\mathbf{I}^*(u,v)$.
This step is applied to all three color channels of $\mathbf{I}(u,v)$.

\paragraph*{Optimal Seam}
On behalf of a smooth transition between consecutive images, it is advantageous to place the image seam where the image difference is small.
For that purpose, an optimal path is computed by dynamic programming.
With two consecutive images $\mathbf{I}_n(u,v)$ and $\mathbf{I}_{n+1}(u,v)$ overlapping along the pipe direction within the regions $\mathbf{I}_n^{ov}$ and $\mathbf{I}_{n+1}^{ov}$, we use the normalized, absolute difference

\begin{equation}
\mathbf{D}(u,v) = \frac{|\mathbf{I}_n^{ov} - \mathbf{I}_{n+1}^{ov}|}{255}
\end{equation}

for the grey-values as error criterion.

To avoid a frayed seam, we add an additional cost to penalize a transition from one possible seam element $s(u,v)$ to one in the next column $s(u,v+1)$ by $\delta = (u_v - u_{v+1})^2$, favoring horizontal cuts through the images.
For every element in the current column with index $v$ the cost for getting there is calculated by

\begin{equation}
\mathbf{w}_{v+1}(u) = \alpha \mathbf{D}(u,v+1) + \min \left( \beta \frac{\delta}{h_{D}} + \mathbf{w}_{v}(u) \right)
\end{equation}

with $h_D$ being the height of the difference image. 
The weights $\alpha$ and $\beta$ control the influence of the difference image and the vertical distance respectively.
With the optimal seam at hand, the transition between two consecutive images takes place where their difference is minimal, while areas with tiny registration errors get excised.

\paragraph*{Poisson Blending}

As final step of the image refinement, we apply Poisson Blending \cite{Perez.2003} along the image seams.
The blending is formulated as an Least-Squares problem, constrained by the previous image at one side and by the current image at the other.
Thereby, the image gradients can be preserved, but leaps are avoided. 
We limit the range of the blending to only a few pixels, to preserve computation time.

\section{Automatic Annotation}
\label{[sec:automaticAnnotation]}

Automatic annotation and damage classification of the enhanced images is treated as a semantic labeling problem which is tackled using deep convolutional neural networks. 
As mentioned in \autoref{sec:intro}, many algorithms rely on image processing approaches and heuristics to detect defects in sewer pipes. 
Although those algorithms are able to detect some defects reliably, a common problem is the relatively low amount and quality of data available. On the one hand, this constrains the number of detectable defect classes, due to the lack of examples per class, whereas on the other hand, the varying visual appearance of those classes makes it nearly impossible to find a single general detection method on that few examples.

We think that given a sufficient amount of high quality labeled data, a deep convolutional neural network can learn to detect and differentiate between a variety of different classes with almost expert-like accuracy.

\subsection{Data Acquisition}
\label{ssec:data}
In order to successfully train deep neural networks, large amounts of data are needed. Additionally in the case of semantic labeling, this data must also be labeled in a pixel-precise way, meaning that each pixel must be assigned to one specific class.

Given the enhanced, unrolled image produced by the method in \autoref{sec:imagePreProcessing} and the expertise of specially trained experts, we were able to produce such pixel-precise labeling for $111$ sewer pipes covering almost $4.6$ kilometers. We selected the pipes to represent a variety of materials ($61$ stoneware and $50$ concrete) and diameters ($200$ to $500$ millimeters). The number of pixels representing the circumference was set to $1200$ and the resulting image resolution was computed accordingly.
Overall, we manually annotated $1200$x$7123693$ pixels of unwrapped pipe images.
Since each single image is too large to be used for training as a whole, we decided to split them into equally sized, overlapping chunks of size $600$x$1200$.

\begin{figure}[h]
	\centering
	\sidesubfloat[]{\includegraphics[width=0.8\linewidth]{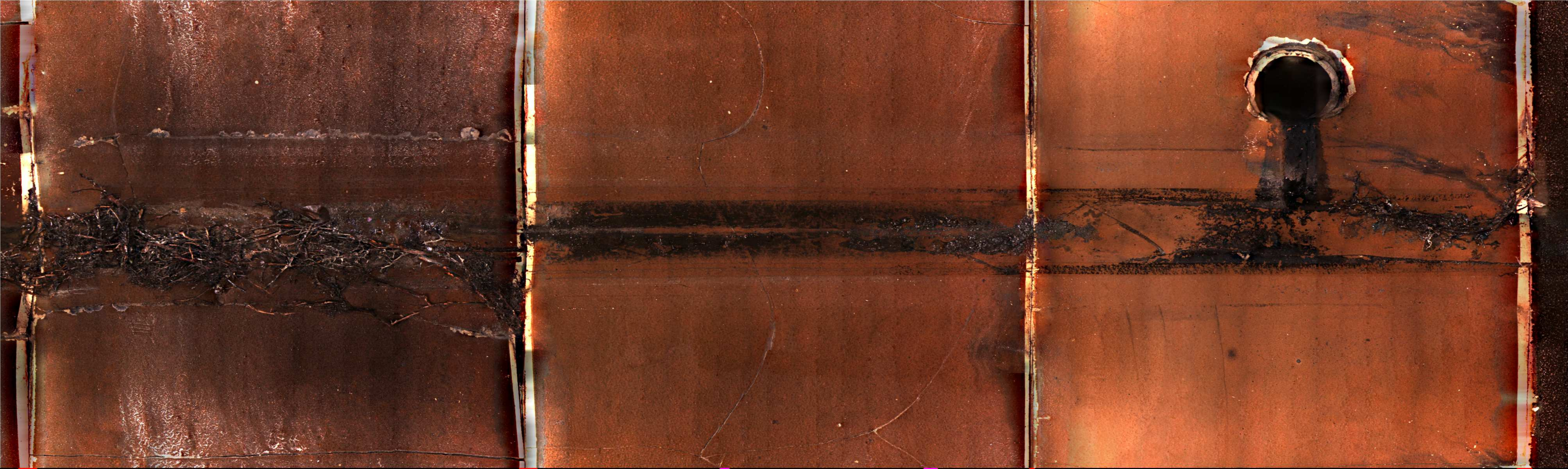}}
	\vspace{3mm}
	\sidesubfloat[]{\includegraphics[width=0.8\linewidth]{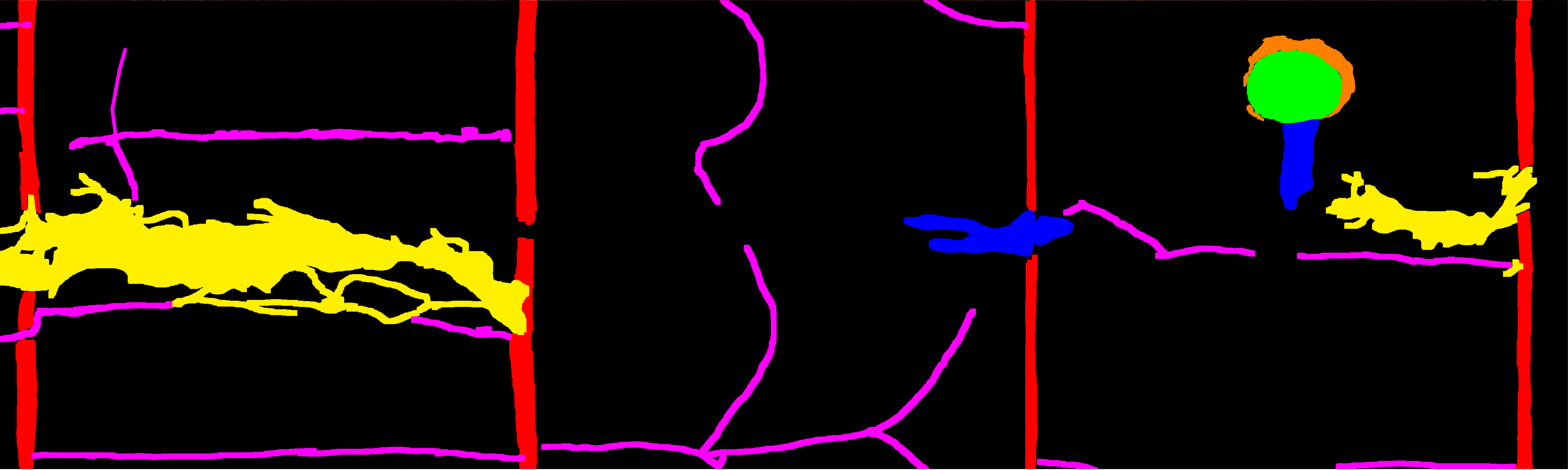}}
	\caption{Pixel-precise labeling (b) of a sewer pipe (a). Classes are: root (yellow), crack (magenta), residue (blue), spalling (orange), connection (green), joint (red).}
	\label{fig:labeling}
\end{figure}

For the annotation of the images, we selected some of the most common defects as well as structural elements. Overall, $9$ classes were used. Regarding defects the classes are \textit{residue, crack, root, obstacle and erosion/spalling}, whereas for the structural elements we used \textit{joint, connection and shaft}.
A labeled example image can be seen in \autoref{fig:labeling}.

\floatsetup[figure]{style=plain,subcapbesideposition=center}

\floatsetup[table]{style=plain,subcapbesideposition=center}
\begin{table}
\begin{tabular}{ l | r | r | r}
	\multicolumn{1}{c|}{Class} & \multicolumn{1}{|c|}{No. objects} & \multicolumn{1}{|c|}{No. pixels} & \multicolumn{1}{|c}{\%}\\
	\hline			
	\hline		
	Background	& -			& 8062984727	& 94.3212\\
	Connection	& 717		& 29318800		& 0.343\\
	Joint		& 5119		& 126881255		& 1.4843\\
	Residue		& 666		& 92288494		& 1.0796\\
	Crack		& 1687		& 11258490		& 0.1317\\
	Root		& 1558		& 45812512		& 0.5359\\
	Obstacle	& 75		& 190887		& 0.0022\\
	Spelling	& 1654		& 94677248		& 1.1075\\
	Shaft		& 148		& 85019187		& 0.9946  
\end{tabular}
\caption{Number of objects and number of pixels within each class. The last column shows the fraction of the class specific pixels to the number of pixels overall.}
\end{table}

\subsection{Network Topology}
The network structure we used to perform semantic labeling is based on the \textit{Full-Resolution Residual Networks (FRRN)} by Pohlen \etal.~\cite{Pohlen2017}. 
In their work, the authors develop a novel topology of deep convolutional neural network aimed at semantic labeling tasks.

\begin{figure}
	\centering
	\includegraphics[width=0.9\linewidth]{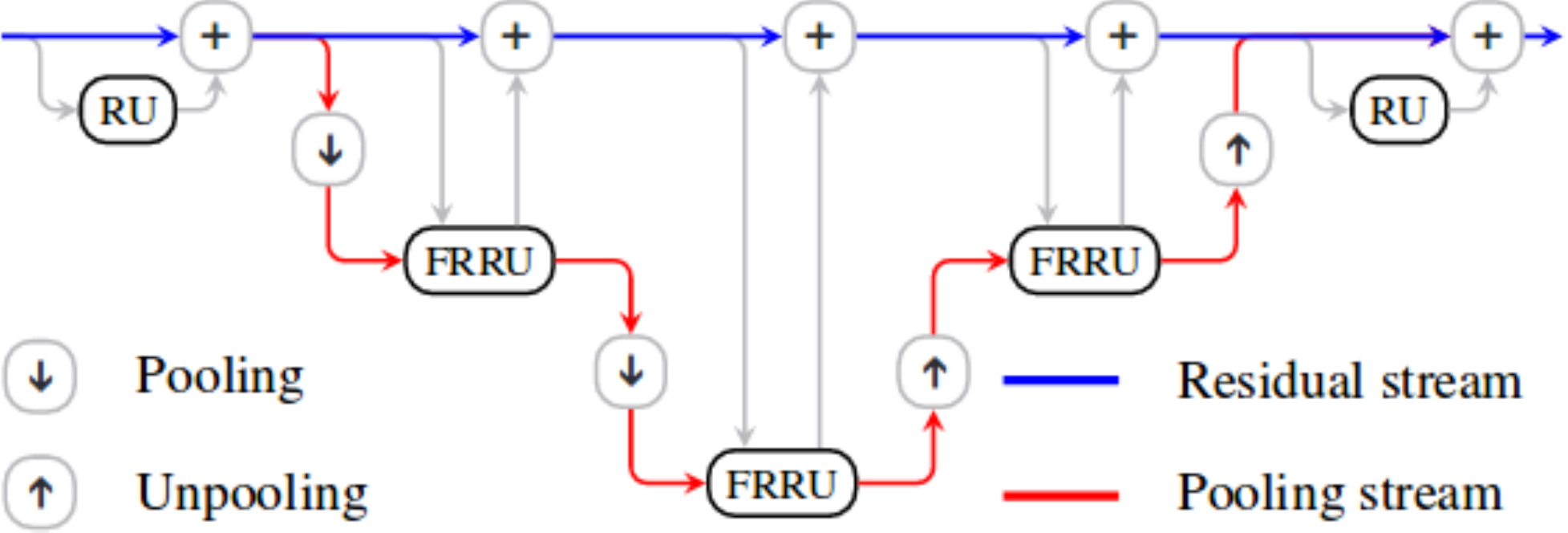}
	\caption{General structure of an FRRN. The recognition stream (red) undergoes down-sampling and feature extraction, whereas the full resolution residual stream (blue) collects all subsequent features. Taken from \cite{Pohlen2017}.}
	\label{fig:frrn}
\end{figure}

The principle idea of this structure is to have two data streams through the whole network. One stream is responsible for object recognition, which undergoes a classical pipeline of feature extraction and down-sampling to learn robust features for object recognition, whereas the other stream is kept at full input resolution to learn features for precise object segmentation. Successively features learned by the recognition (pooling) stream are up-sampled and fused into the segmentation (residual) stream. This way, even more complex features are generated for the final pixel-wise labeling. The general structure of an FRRN can be seen in \autoref{fig:frrn}.
We adapted the original FRRN structure to better fit our problem and to reduce complexity, as well as computation time and model size. Compared to FRRN\textsubscript{A} structure in \cite{Pohlen2017}, we changed the number of full resolution residual units (FRRU) per resolution level to 3 and the number of filters to $24$ for the pooling stream and $16$ for the residual stream.

\begin{figure*}[tb]
\centering
\subfloat[unwrapped pipe surface generated with a commercial sewer pipe inspection\label{subfig:comparsion_a}]{\includegraphics[width=0.7\linewidth,height=3cm]{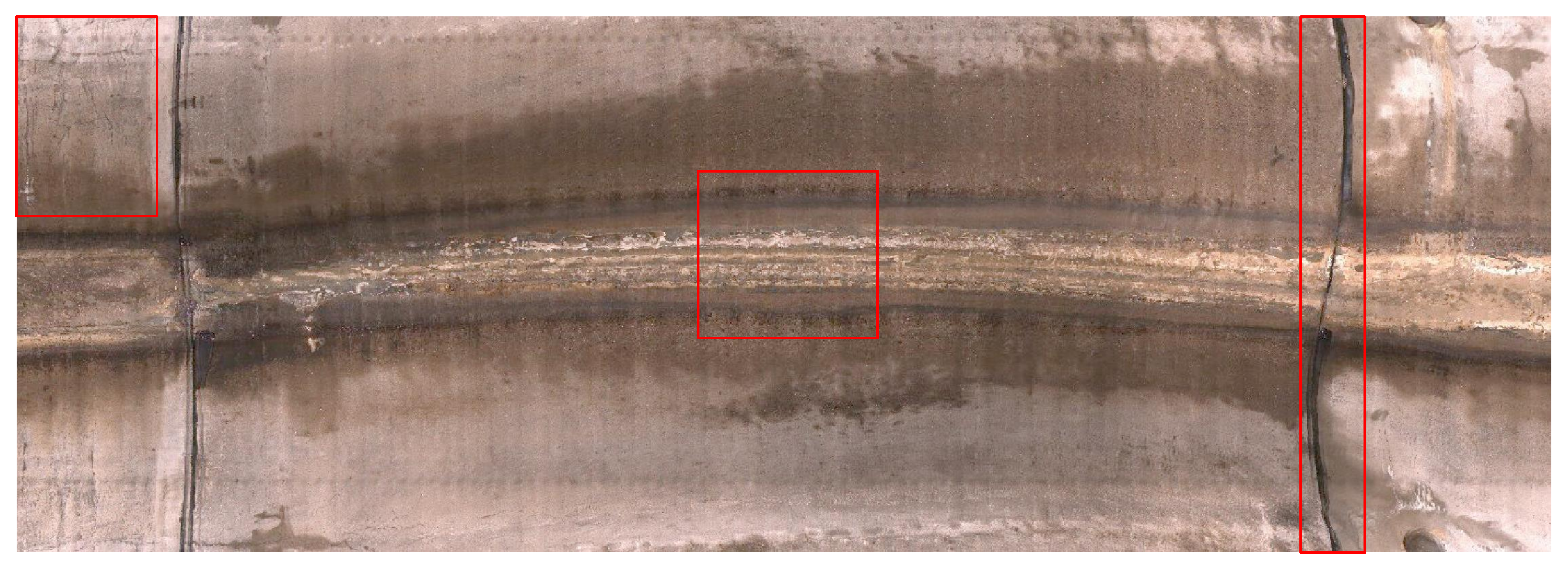}}
\subfloat[camera path on the pipe axis\label{subfig:comparsion_b}]{\includegraphics[width=0.28\linewidth,height=3cm]{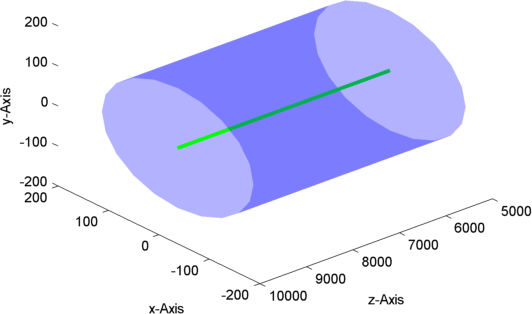}}
\qquad
\subfloat[unwrapped pipe surface generated with our system\label{subfig:comparsion_c}]{\includegraphics[width=0.7\linewidth,height=3cm]{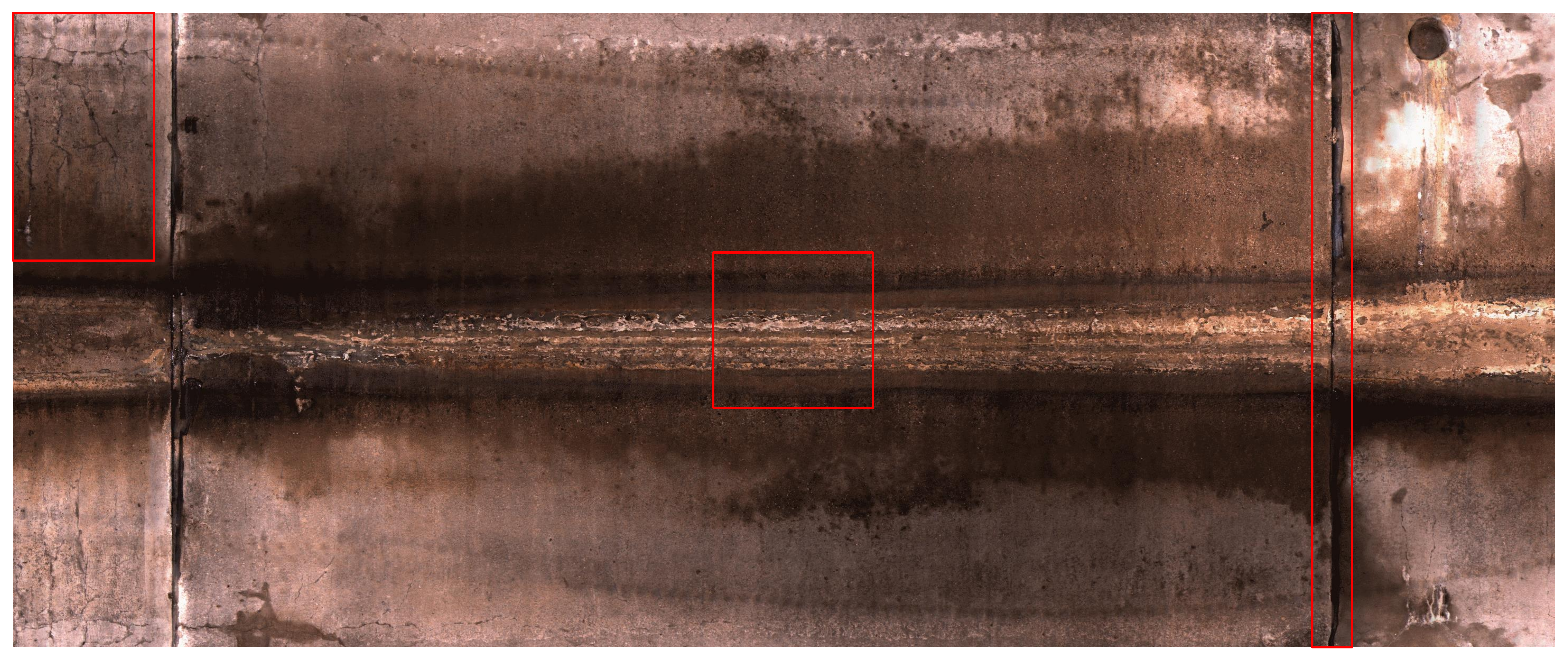}}
\subfloat[estimated camera motion path\label{subfig:comparsion_d}]{\includegraphics[width=0.28\linewidth,height=3cm]{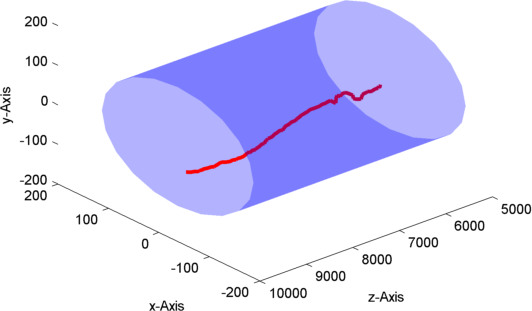}}
\caption{Parts of an unwrapped pipe surface with associated camera motion paths used for back-projection. Top: commercial sewer pipe inspection software, which assumes a camera path along the pipe axis. Bottom: same section processed with our system. The camera path was estimated as described, resulting in a motion-artifact-free pipe unwrap.}
\label{fig:comparsion}
\end{figure*}

\subsection{Training}
For training, the data generated in \autoref{ssec:data} was split into a training and a test set using a ratio of $80$:$20$. 
Due to the large image size, training was performed on downscaled versions of the data with a size of $256$x$512$. 
This enabled us to decrease training time at no cost in terms of quality compared to the original resolution. 
Furthermore, the images were converted to \textit{YCbCr} space and contrast normalized in a windowed fashion to compensate for the varying materials and color (e.g.~of residues) and bright reflections due to wet surfaces.

To further increase the amount of training data available, minor data augmentation was applied. Since sewer pipe inspection is direction independent and the pipes are symmetrical, we randomly flipped the input images either vertically or horizontally with a probability of $\frac{1}{3}$.

Training was performed on a single \textit{Nvidia Titan X} for $72$ hours using \textit{Tensorflow}. We optimized a bootstrapped cross entropy as introduced in \cite{Wu2016}. The idea is to only take a certain percentage of pixels $p$ into account, which are misclassified or correctly classified with a low class probability
\begin{equation}
l = -\frac{1}{K}\sum_{i=1}^{N}1[p_{i,y_i} < t_K]\log p_{i,y_i}
\end{equation}
where $p_{i,y_i}$ is the posterior probability for image pixel $i$ and its corresponding target class $y_i$ and $t_K$ is a threshold chosen so that $K$ elements fall below this. We selected $K = N \cdot p$ with $p=0.1$ where $N$ is the total number of image pixels.

\begin{figure}
	\centering
	\includegraphics[width=0.9\linewidth,height=5cm]{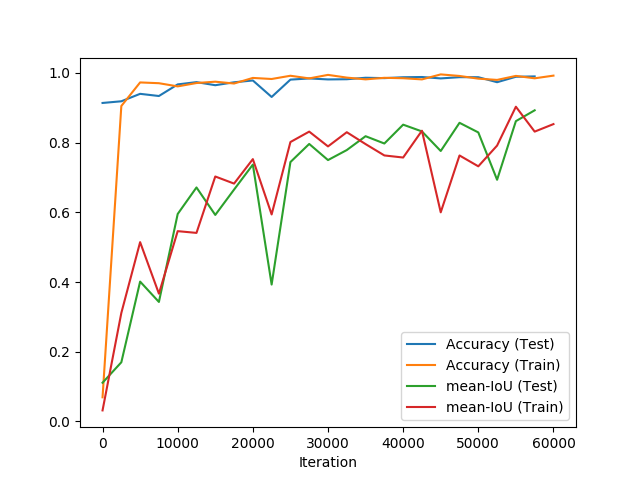}
	\caption{Evolution of accuracy an mean-IoU over training iterations on the training and validation sets.}
	\label{fig:testPlot}
\end{figure}

Training was performed using the Adam optimizer \cite{Kingma2014} with a constant learning rate of $10^{-4}$. In \autoref{fig:testPlot}, the evolution of accuracy and mean-IoU can be seen. As expected, the accuracy increases rapidly because most of the images contain a large portion of background.

\section{Results}

In this section, we first show results of our system for the generation of images depicting unwrapped pipe surfaces.
We compare the results to images produced by a commercial sewer pipe inspection software.
In the second part, we present the results and benchmarks for our damage detection algorithm.

\subsection{Results -- Unrolled Pipes}

\begin{figure}[h]
\centering
\subfloat{\includegraphics[angle=90,width=1.0\linewidth]{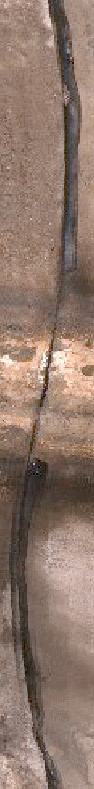}}
\vspace{3mm}
\subfloat{\includegraphics[angle=90,width=1.0\linewidth]{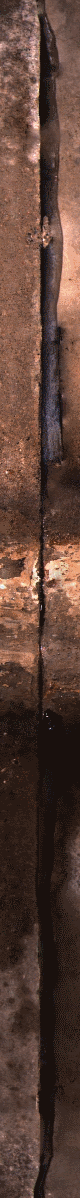}}
\caption{Detail views for the pipe couplings (cropped and rotated from \autoref{fig:comparsion}). \textbf{Top:} Typical artifact due to an incorrect camera pose estimation by the commercial sewer pipe inspection software. \textbf{Bottom:} The distortion free pipe coupling as a result of the camera pose estimation.}
\label{fig:comparisonConnections}
\end{figure}

The images of \autoref{fig:comparsion} illustrate the influence of the motion path estimation.
A pipe section is shown on the left side, generated with the commercial sewer pipe inspection software in \autoref{subfig:comparsion_a}, and with our system in \autoref{subfig:comparsion_c}.
The images on the right show the camera motion paths used for the calculation of the back projection.
As depicted in \autoref{subfig:comparsion_b}, a camera motion along the pipe axis was assumed for the generation \autoref{subfig:comparsion_a}.
Therefore, the actual camera movement causes several artifacts in the final image.
It creates foremost the illusion of a bent pipe.
A not centrically placed camera is also easily to notice by the depicted wavy pipe couplings.
In \autoref{fig:comparisonConnections}, details of the pipe couplings are shown for comparison.

\begin{figure}[h]
\centering
\subfloat{\includegraphics[height=3.1cm, width=3.1cm]{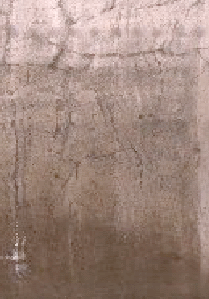}}
\hspace{3mm}
\subfloat{\includegraphics[height=3.1cm, width=3.1cm]{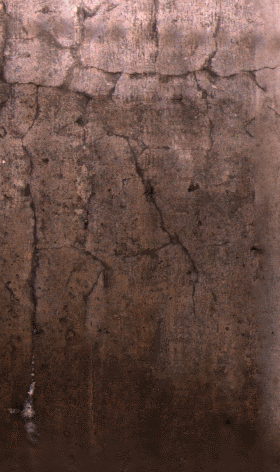}}
\caption{Detail views of cracks cropped from \autoref{fig:comparsion}. \textbf{Left:} Due to small registration errors some cracks appear twice and blurry. \textbf{Right:} The cracks are correctly registered due to camera pose and optimal path estimation.}
\label{fig:comparisonCracks}
\end{figure}

The images in \autoref{fig:comparisonCracks} show close up views of a pipe surface region with many cracks.
In the left image, some cracks appear twice showing ghosting effects due to the lack of camera pose information.
With our estimation of the camera positions, these artifacts disappear (right).
Furthermore, small registration errors get excised by the estimation of the optimal path.
In addition, our algorithm led to a noticeable improvement of the image resolution and contrast.

\begin{figure}[ht]
\centering
\subfloat{\includegraphics[width=0.46\linewidth,height=3cm]{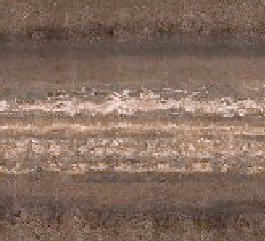}}
\hspace{3mm}
\subfloat{\includegraphics[width=0.46\linewidth,height=3cm]{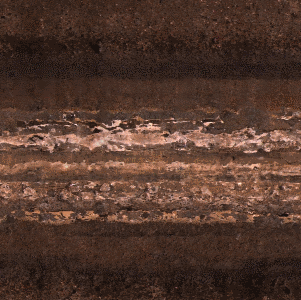}}
\caption{Detail views of the pipe bottom cropped from \autoref{fig:comparsion}. \textbf{Left:} Result of the commercial system with strong blurring and limited contrast. \textbf{Right:} The result of our system with enhanced details and image contrast.}
\label{fig:comparisonBottom}
\end{figure}

These changes can be also noticed in the right image of \autoref{fig:comparisonBottom} where depositions at the pipe bottom can be inspected in much more detail.
In the left image of \autoref{fig:comparisonBottom}, produced by the commercial system, these details got lost due to the low resolution.

\floatsetup[table]{style=plain,subcapbesideposition=center}
\begin{table*}
	\begin{tabular}{l||c|c|c|c|c|c|c|c|c||c}
\multicolumn{1}{c||}{Class} & \multicolumn{9}{c||}{Confusion}& mean-IoU		\\\hline\hline
Background & \textbf{99.52} & 0.11 & 0.01 & 0.18 & 0.05 & 0.04 & 0.00 & 0.08 & 0.01 & 0.988\\
Joint & 18.35 & \textbf{80.91} & 0.00 & 0.18 & 0.15 & 0.07 & 0.01 & 0.27 & 0.05 & 0.752\\
Connection & 2.86 & 0.01 & \textbf{95.80} & 0.56 & 0.10 & 0.17 & 0.00 & 0.48 & 0.02 & 0.917\\
Residue & 4.33 & 0.15 & 0.01 & \textbf{95.31} & 0.01 & 0.09 & 0.01 & 0.09 & 0.01 & 0.879\\
Crack & 28.32 & 0.89 & 0.38 & 0.18 & \textbf{70.03} & 0.09 & 0.00 & 0.10 & 0.00 & 0.584\\
Root & 9.43 & 1.28 & 0.04 & 0.40 & 0.32 & \textbf{88.49} & 0.00 & 0.03 & 0.01 & 0.794\\
Obstacle & 34.14 & 17.40 & 0.00 & 3.70 & 0.53 & 1.70 & \textbf{42.29} & 0.24 & 0.00 & 0.374\\
Spelling & 6.30 & 0.37 & 0.26 & 0.09 & 0.03 & 0.02 & 0.00 & \textbf{92.89} & 0.03 & 0.894\\
Shaft & 3.98 & 0.01 & 0.22 & 0.42 & 0.00 & 0.01 & 0.00 & 0.06 & \textbf{95.30} & 0.949
	\end{tabular}
	\caption{Confusion matrix representing accuracy on a pixel level on the test set. The last column represents the mean-IoU per class averaged over all images.}
	\label{tab:confusion}
\end{table*}

\subsection{Results -- Automatic Annotation}
\label{ssec:resultsAnnot}

\begin{figure}[ht!]
	\centering
	\subfloat[Input]{\includegraphics[width=0.3\linewidth,height=3.8cm]{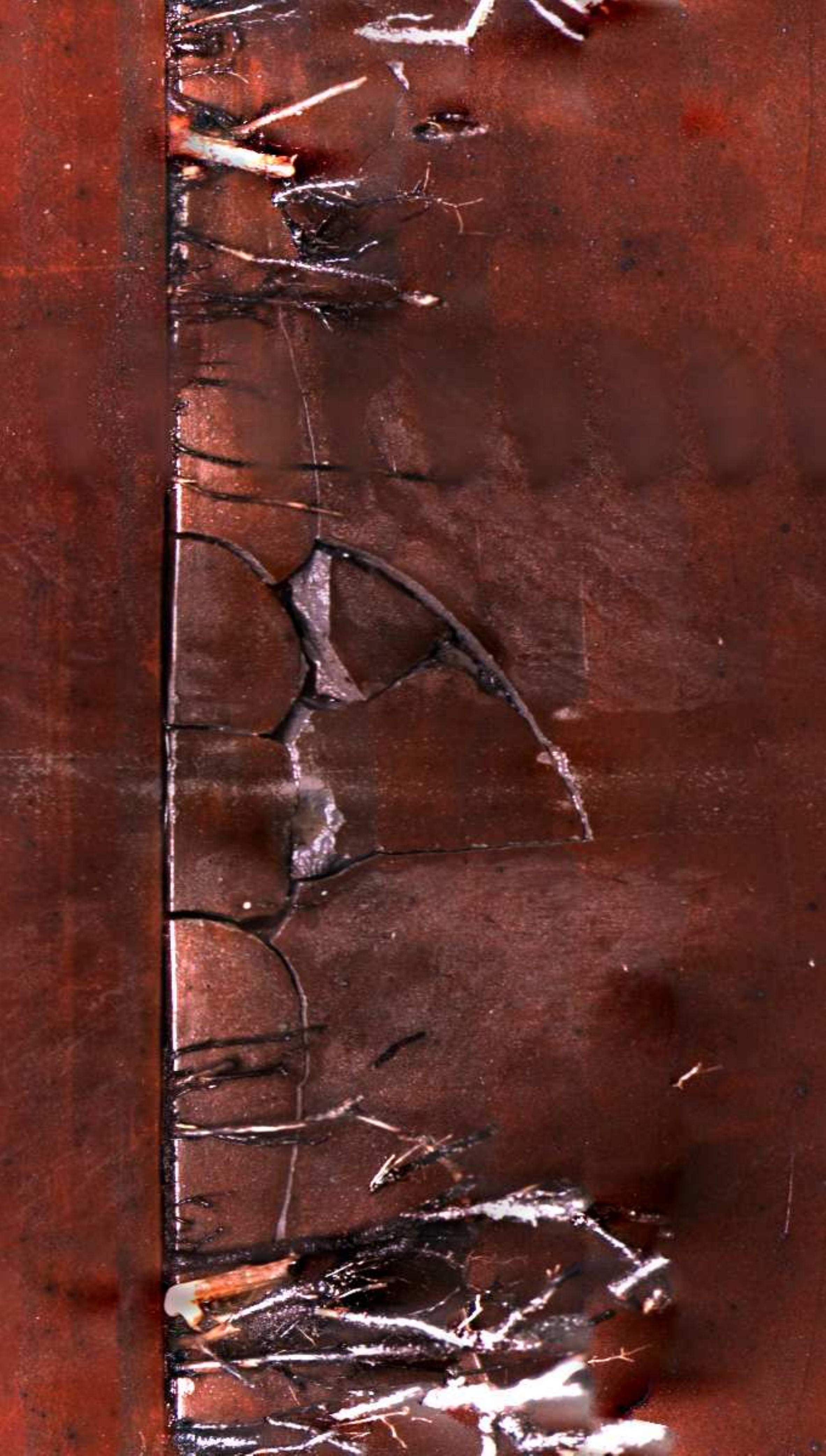}}
	\vspace{1mm}
	\subfloat[Ground Truth]{\includegraphics[width=0.3\linewidth,height=3.8cm]{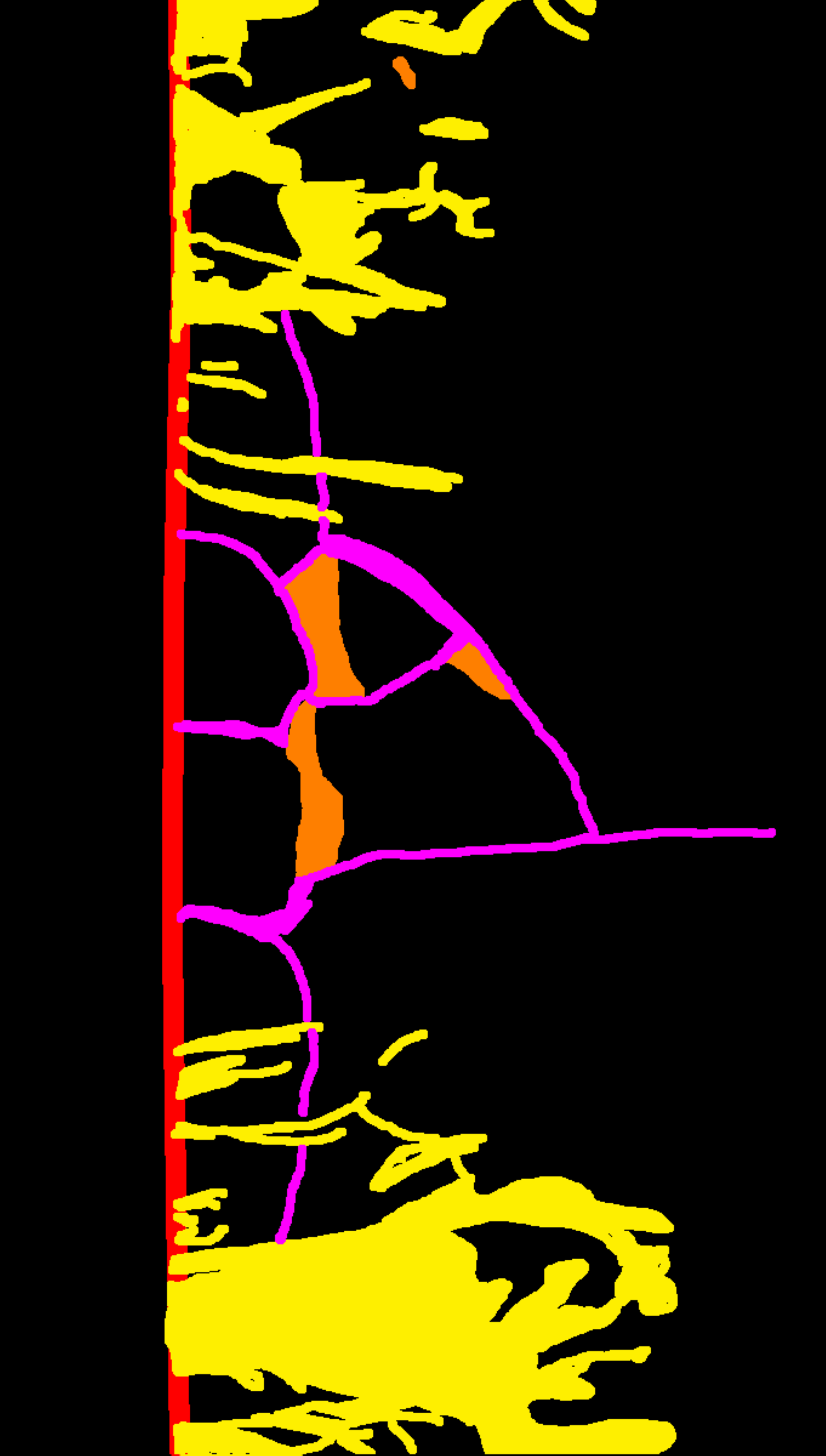}}
	\vspace{1mm}
	\subfloat[Prediction]{\includegraphics[width=0.3\linewidth,height=3.8cm]{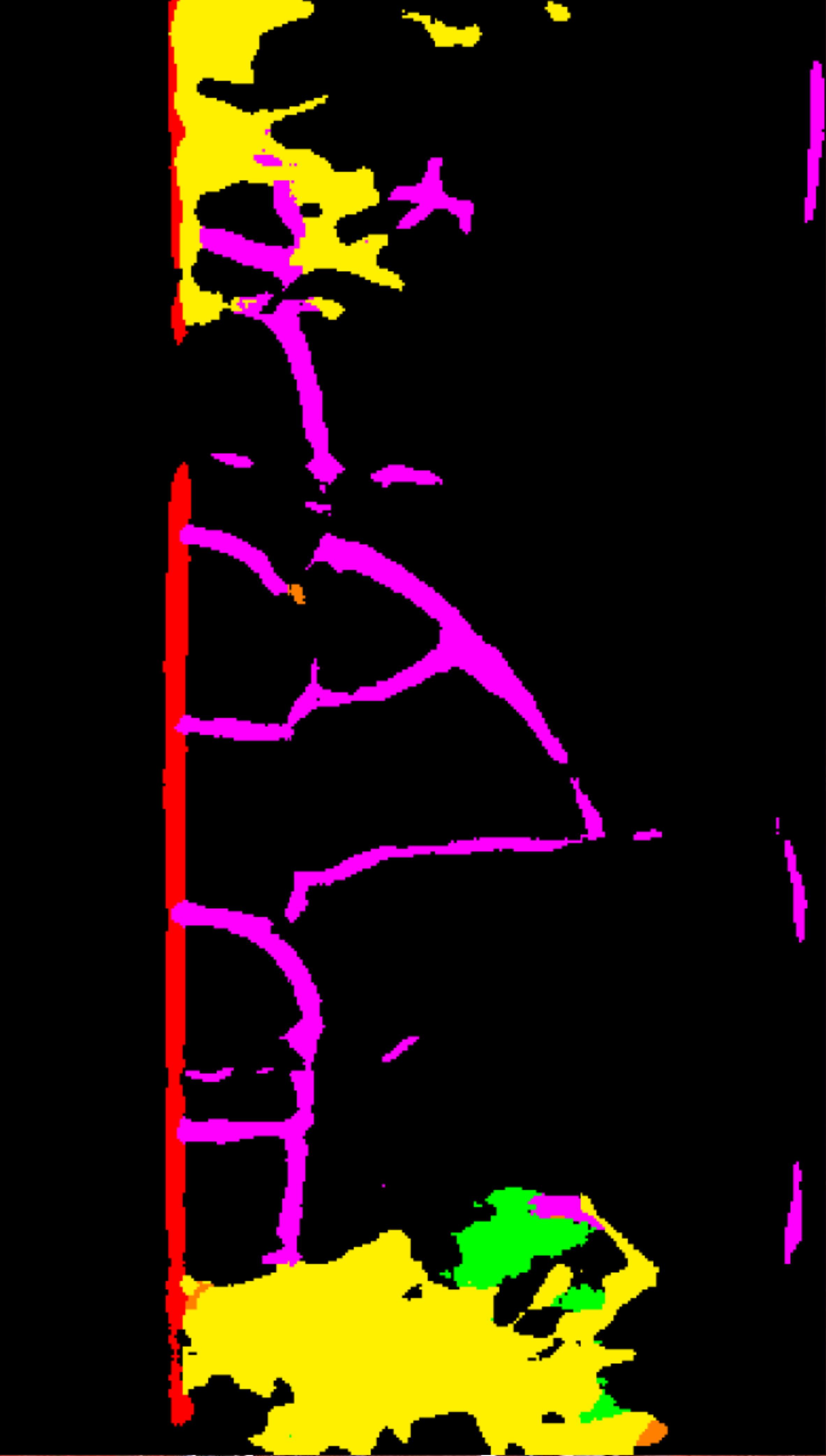}}
	\caption{Labeling of a stoneware pipe. Joints, cracks and roots are reasonably detected, whereas spalling is missed. Also some dark areas are confused with joints.}
	\label{fig:resultsVeryGood}
\end{figure}

\begin{figure}[ht!]
	\centering
	\subfloat[Input]{\includegraphics[width=0.3\linewidth,height=3.2cm]{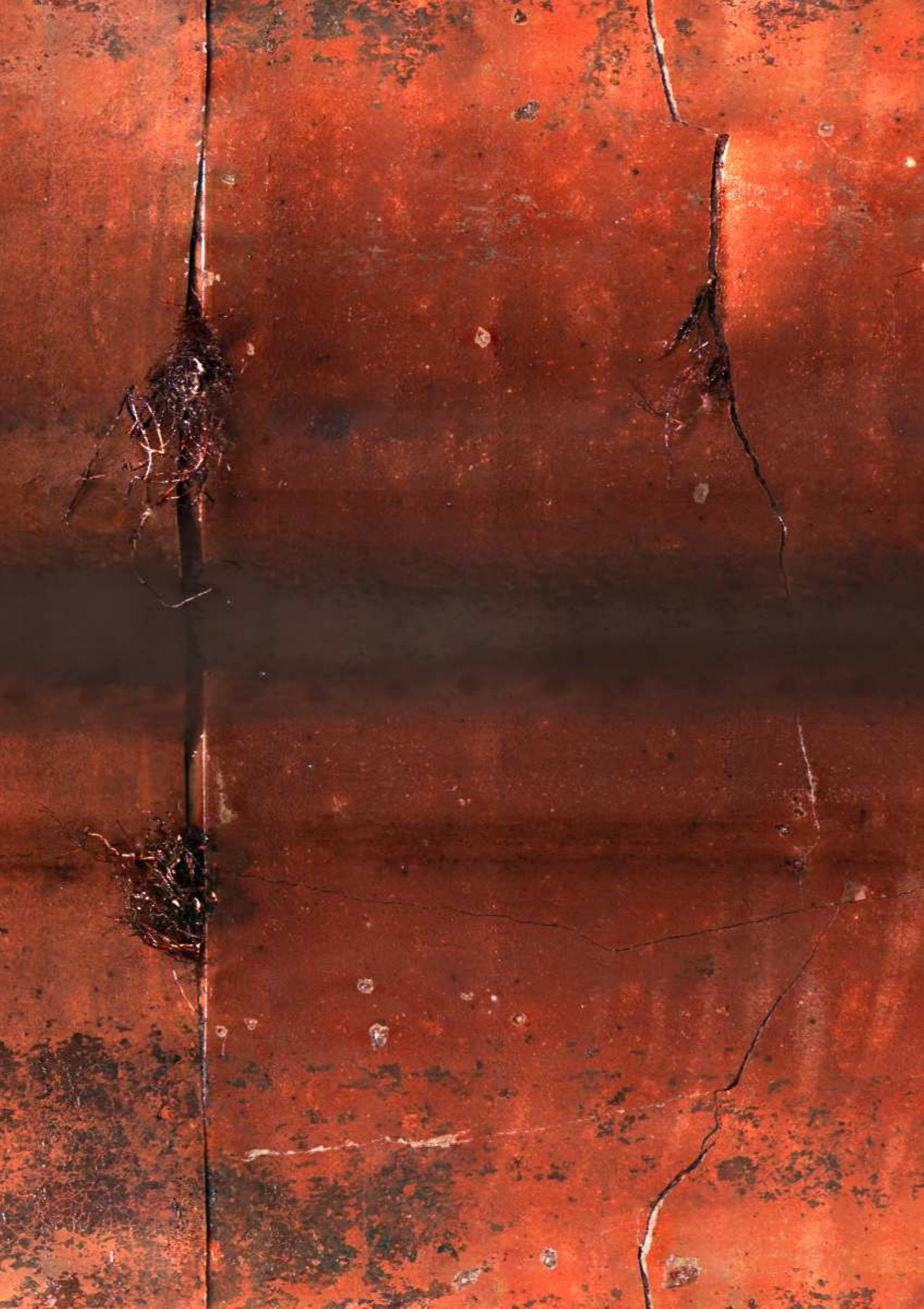}}
	\vspace{1mm}
	\subfloat[Ground Truth]{\includegraphics[width=0.3\linewidth,height=3.2cm]{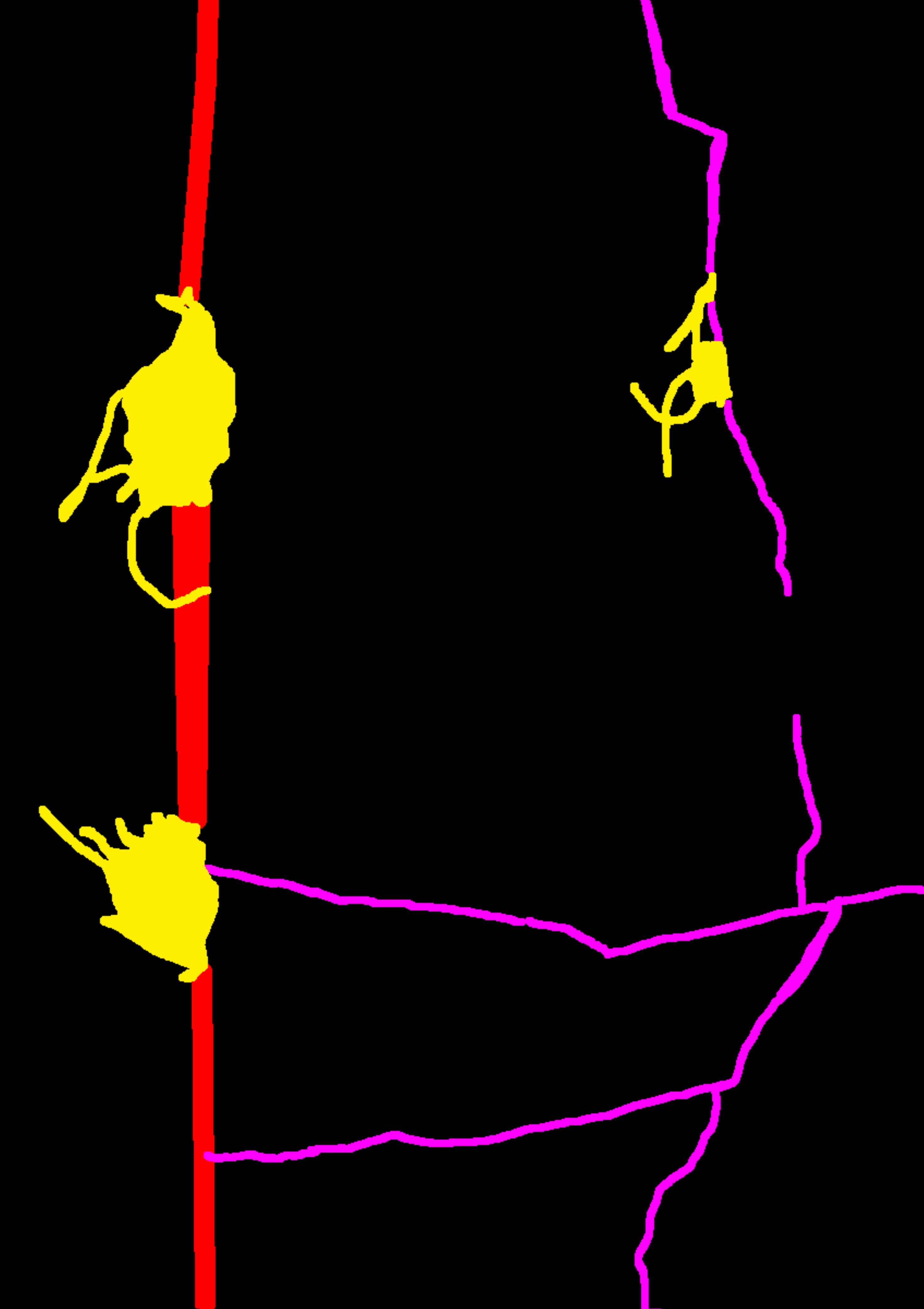}}
	\vspace{1mm}
	\subfloat[Prediciton]{\includegraphics[width=0.3\linewidth,height=3.2cm]{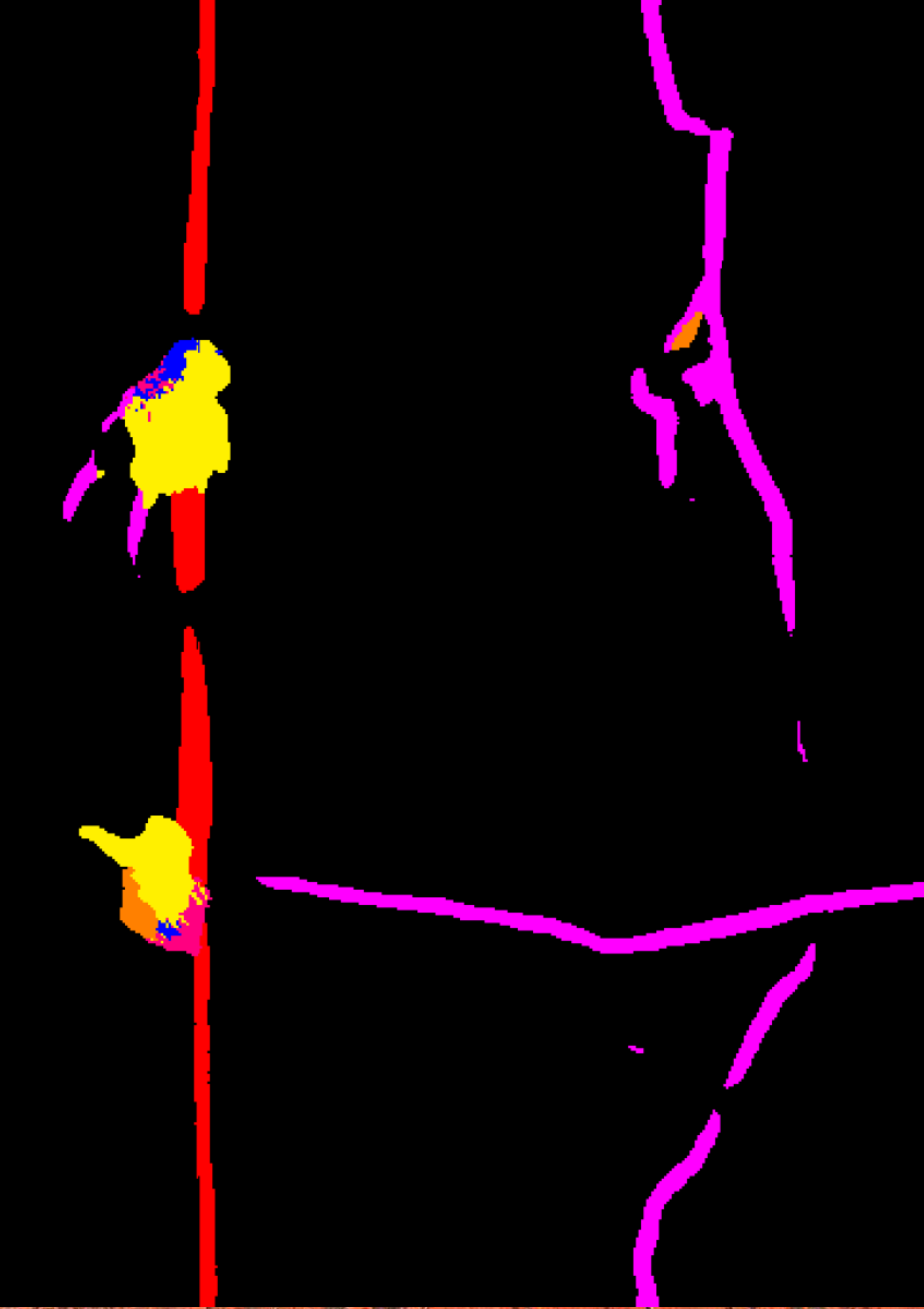}}
	\caption{Labeling of a stoneware pipe. Cracks are generally easily detected. Still some fine roots are mistaken for cracks due to their dark color. Connections are generally no problem.}
	\label{fig:resultsGood}
\end{figure}

\begin{figure}[ht!]
	\centering
	\subfloat[Input]{\includegraphics[width=0.3\linewidth,height=3.5cm]{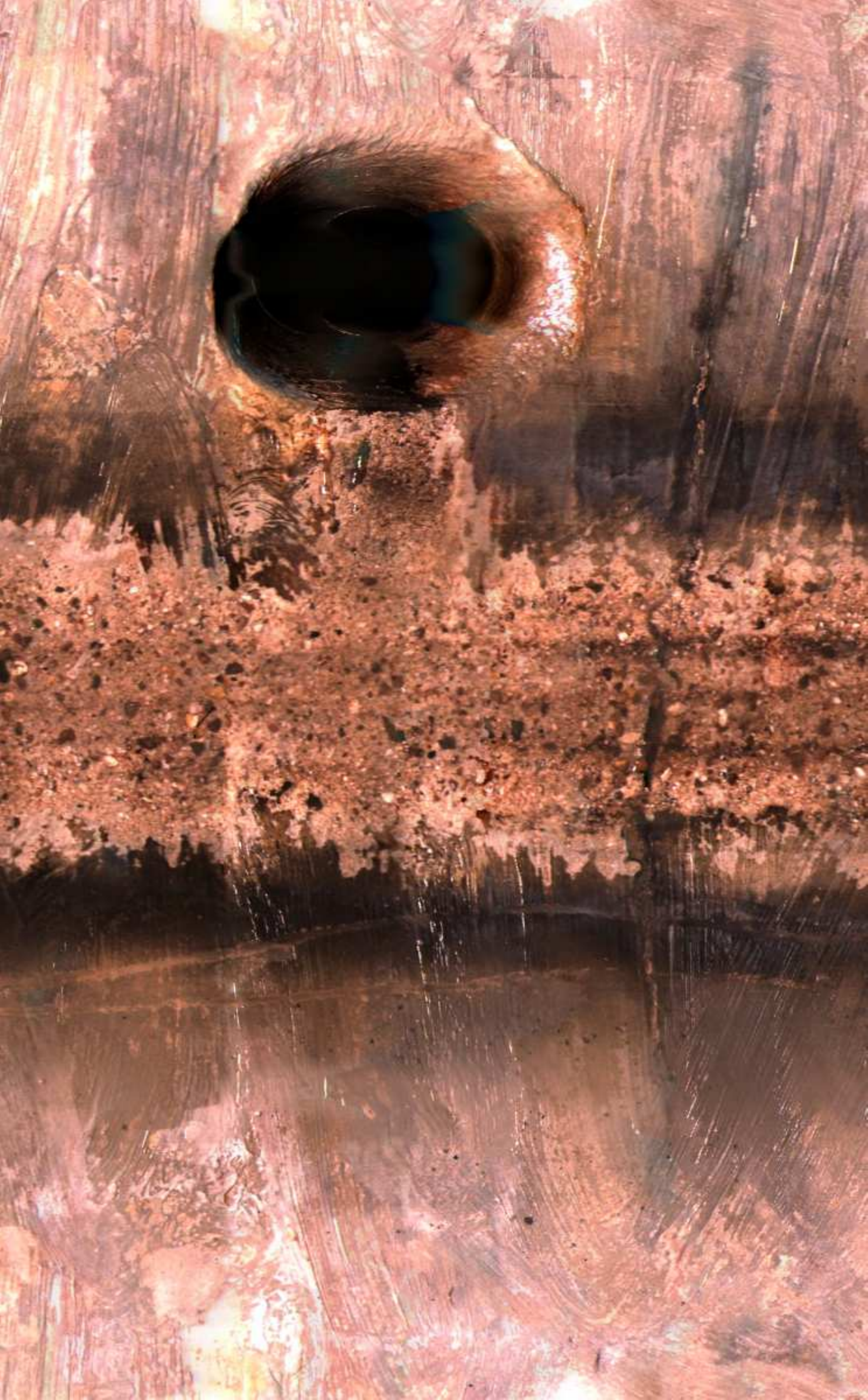}}
	\vspace{1mm}
	\subfloat[Ground Truth]{\includegraphics[width=0.3\linewidth,height=3.5cm]{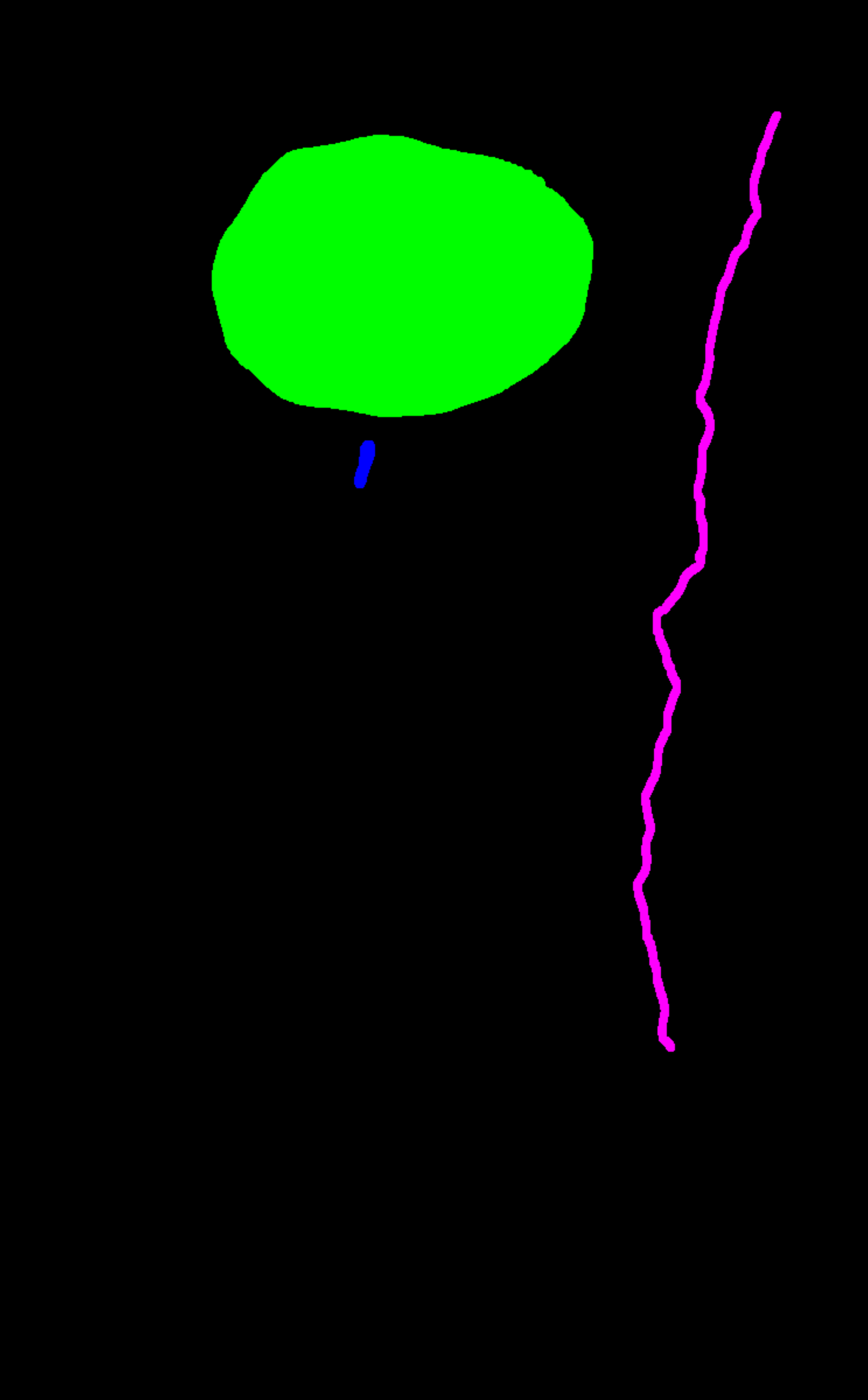}}
	\vspace{1mm}	
	\subfloat[Prediction]{\includegraphics[width=0.3\linewidth,height=3.5cm]{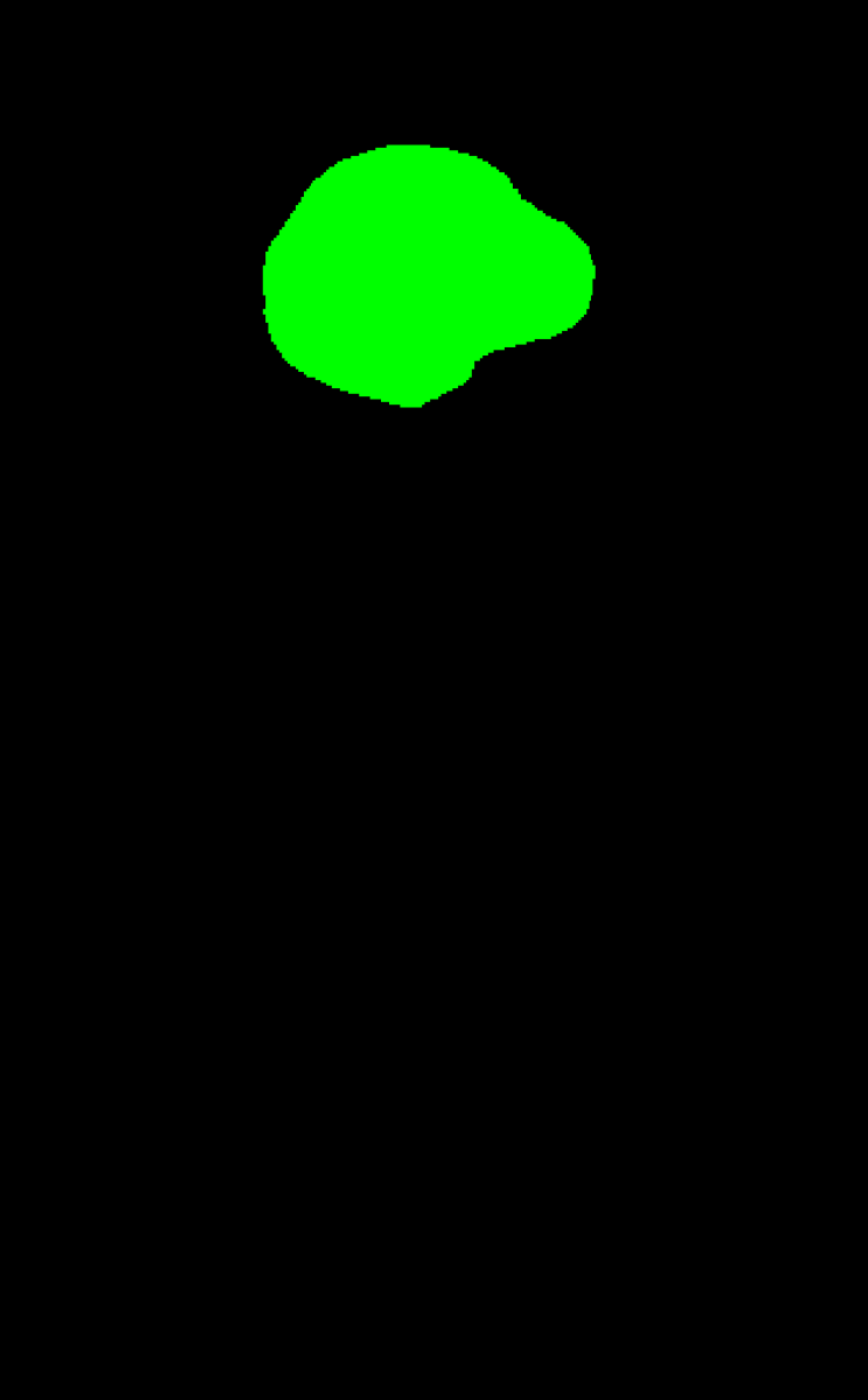}}
	\caption{Labeling of a concrete pipe. Due to the high roughness of the surface, cracks can easily be missed.}
	\label{fig:resultsMissedCrack}	
\end{figure}

In \autoref{fig:resultsVeryGood}, \autoref{fig:resultsGood} and \autoref{fig:resultsMissedCrack}, some exemplary labelings produced by the presented system are shown. It can be seen that structural elements can be detected and classified reliably, regardless of the pipes material. For defects however, there are some differences among the different pipe types. Depending on the material of the pipe, some defects are more likely to be missed. In \autoref{fig:resultsVeryGood}, and \autoref{fig:resultsGood}, cracks and roots are detected reasonably and classified as such, whereas in \autoref{fig:resultsMissedCrack} the crack is missed completely. Also, sometimes cracks and roots, although detected, are mistaken for each other due to their sometimes similar color. All in all, our system produces visually satisfying results.

\autoref{tab:confusion} shows the confusion matrix and mean-IoU on the test set and gives a more detailed overview of the results. It can be seen that the most problematic class is \textit{obstacle} (seventh line), which most often gets confused with the background. This makes sense in that usually an obstacle that is viewed from the center of the pipe like on the unwrapped images is virtually invisible and cannot be distinguished from the background. Only in cases where there are also color changes, the obstacle can be detected.
The second most problematic class is \textit{crack} (fifth line) which also gets missed often, due to the complex texture of the concrete pipes. An example can be seen in \autoref{fig:resultsMissedCrack}.

In general it seems that most often when a defect or structure is missed, the problem is not that it gets misclassified but rather is missed at all and mistaken for background. This could be a serious problem in terms of risk assessment, which we look out to have a closer look on in future work (see \autoref{sec:outlook}).

\section{Outlook}
\label{sec:outlook}
Until now, we have only used a state-of-the-art network topology that shows general good performance in semantic segmentation tasks. We plan to enhance the used structure to incorporate the shape of a pipe in the sense that the images wrap around and that the first and last row are correlated. This way we hope to reduce the number of parameters needed, while keeping the quality of the results. This in turn would reduce computation time and model size.
Furthermore, we aim at increasing the resolution used for training and prediction to overcome the problem of missing thin cracks due to down-sampling.

Although mean-IoU is a good measure to get an intuition for the quality of the system, it is in a sense very academic.
In terms of risk assessment and sanitation planning, detection rate and false positive rate are perhaps the most interesting measures.
We plan to also evaluate our results based on those measures and to give a qualitative evaluation (e.g.~in terms of risk) with the help of experts.

Regarding the image enhancement side of this work, there are two possible ways to go. On the one hand we could use greatly improved images, which would lead to track-able features. On the other hand, one can use stereo optical techniques to retrieve a 3d model of the pipe. In addition to the used defects mentioned in \autoref{ssec:data}, this would enable us to also detect open joints which can be seen as high risk defects due to their influence on the pipe's structural integrity. Furthermore, it would likely improve the detection of obstacles for the reasons mentioned in \autoref{ssec:resultsAnnot}.

\section{Conclusion}
In this paper, we present a method for enhancing low quality fisheye images of sewer pipes and produce high quality unwraps that then are used for automatic detection and classification of defects and structural elements. We show that, given a sufficient amount of data, a single system can be capable of detecting a wide range of defects despite the large visual variations. Although  there are still some problems to tackle, the system achieves good results in terms of accuracy and IoU.

{\small
\bibliographystyle{ieee}
\bibliography{ms}

\begin{thebibliography}{10}\itemsep=-1pt

\bibitem{Cooper.1998}
D.~Cooper, T.~P. Pridmore, and N.~Taylor.
\newblock Towards the recovery of extrinsic camera parameters from video
  records of sewer surveys.
\newblock {\em Machine Vision and Applications}, 11(2):53--63, 1998.

\bibitem{Esquivel.2005}
S.~Esquivel, R.~Koch, and H.~Rehse.
\newblock Reconstruction of sewer shaft profiles from fisheye-lens camera
  images.
\newblock In {\em Proc. DAGM}, pages 332--341, 2009.

\bibitem{Esquivel.2010}
S.~Esquivel, R.~Koch, and H.~Rehse.
\newblock Time budget evaluation for image-based reconstruction of sewer
  shafts.
\newblock In {\em Proc. SPIE 77240, Real-Time Image and Video Processing},
  2010.

\bibitem{Furch.2013}
J.~Furch and P.~Eisert.
\newblock An iterative method for improving feature matches.
\newblock In {\em 2013 International Conference on 3D Vision}, pages 406--413.
  IEEE, 2013.

\bibitem{Huynh2015}
P.~Huynh, R.~Ross, A.~Martchenko, and J.~Devlin.
\newblock Dou-edge evaluation algorithm for automatic thin crack detection in
  pipelines.
\newblock In {\em 2015 IEEE International Conference on Signal and Image
  Processing Applications (ICSIPA)}, pages 191--196, Oct 2015.

\bibitem{Kannala.2008}
J.~Kannala, S.~S. Brandt, and J.~Heikkil{\"a}.
\newblock Measuring and modelling sewer pipes from video.
\newblock {\em Machine Vision and Applications}, 19(2):73--83, 2008.

\bibitem{Kingma2014}
D.~P. Kingma and J.~Ba.
\newblock Adam: {A} method for stochastic optimization.
\newblock {\em CoRR}, abs/1412.6980, 2014.

\bibitem{Perez.2003}
P.~P{\'e}rez, M.~Gangnet, and A.~Blake.
\newblock Poisson image editing.
\newblock In {\em ACM SIGGRAPH 2003 Papers}, pages 313--318, 2003.

\bibitem{Pohlen2017}
T.~Pohlen, A.~Hermans, M.~Mathias, and B.~Leibe.
\newblock Full-resolution residual networks for semantic segmentation in street
  scenes.
\newblock In {\em Proc. IEEE Conf. on Computer Vision and Pattern Recognition
  (CVPR)}, 2017.

\bibitem{Snavely.2005}
N.~Snavely, S.~M. Seitz, and R.~Szeliski.
\newblock Photo tourism: Exploring photo collections in 3d.
\newblock In {\em ACM Siggraph}, pages 835--846, 2006.

\bibitem{Su2015}
T.-C. Su.
\newblock Segmentation of crack and open joint in sewer pipelines based on cctv
  inspection images.
\newblock In {\em Proc. AASRI Int. Conf. on Circuits and Systems}, May 2015.

\bibitem{Wu.2013}
C.~Wu.
\newblock Towards linear-time incremental structure from motion.
\newblock In {\em Proc. Int. Conf. on 3D Vision (3DV)}, pages 127--134, 2013.

\bibitem{Wu.2011}
C.~Wu, S.~Agarwal, B.~Curless, and S.~M. Seitz.
\newblock Multicore bundle adjustment.
\newblock In {\em Proc. Int. Conf. on Computer Vision and Pattern Recognition
  (CVPR)}, pages 3057--3064, 2011.

\bibitem{Wu2015}
W.~Wu, Z.~Liu, and Y.~He.
\newblock Classification of defects with ensemble methods in the automated
  visual inspection of sewer pipes.
\newblock {\em Pattern Analysis and Applications}, 18(2):263--276, 2015.

\bibitem{Wu2016}
Z.~Wu, C.~Shen, and A.~van~den Hengel.
\newblock Bridging category-level and instance-level semantic image
  segmentation.
\newblock {\em CoRR}, abs/1605.06885, 2016.

\bibitem{Yang2008}
M.-D. Yang and T.-C. Su.
\newblock Automated diagnosis of sewer pipe defects based on machine learning
  approaches.
\newblock {\em Expert Systems with Applications}, 35(3):1327 -- 1337, 2008.

\bibitem{Zhang.2011}
Y.~Zhang, R.~Hartley, J.~Mashford, L.~Wang, and S.~Burn.
\newblock Pipeline reconstruction from fisheye images.
\newblock {\em Journal of WSCG}, 2011.

\end{thebibliography}
}

\end{document}